\documentclass{article}
\usepackage{spconf,amsmath,graphicx}
\usepackage{xcolor}
\usepackage{multirow}
\usepackage{hyperref}

\usepackage{enumitem}
\setlist{nosep, leftmargin=14pt}

\usepackage{mwe} % to get dummy images

\title{MV-S\MakeLowercase{win}-T: Mammogram Classification with Multi-View Swin Transformer}
\name{Sushmita Sarker$^*$, Prithul Sarker$^*$, George Bebis, and Alireza Tavakkoli\thanks{$^*$ Equal contribution}}
\address{Department of Computer Science and Engineering, University of Nevada, Reno, USA}
\begin{document}
%\ninept
%
\maketitle
\begin{abstract}

Traditional deep learning approaches for breast cancer classification has predominantly concentrated on single-view analysis. In clinical practice, however, radiologists concurrently examine all views within a mammography exam, leveraging the inherent correlations in these views to effectively detect tumors. Acknowledging the significance of multi-view analysis, some studies have introduced methods that independently process mammogram views, either through distinct convolutional branches or simple fusion strategies, inadvertently leading to a loss of crucial inter-view correlations. 
% \textcolor{red}{Recent endeavors to incorporate attention mechanisms into multi-view mammography analysis have exhibited promise.}
% \textcolor{red}{Nevertheless, these approaches often depend partially on convolution, limiting the complete exploitation of attention mechanisms.} 
In this paper, we propose an innovative multi-view network exclusively based on transformers to address challenges in mammographic image classification. Our approach introduces a novel shifted window-based dynamic attention block, facilitating the effective integration of multi-view information and promoting the coherent transfer of this information between views at the spatial feature map level. Furthermore, we conduct a comprehensive comparative analysis of the performance and effectiveness of transformer-based models under diverse settings, employing the CBIS-DDSM and Vin-Dr Mammo datasets. 
Our code is publicly available at \url{https://github.com/prithuls/MV-Swin-T}
% Our code is publicly available at \href{https://github.com/prithuls/MV-Swin-T}{https://github.com/prithuls/MV-Swin-T}
\end{abstract}
\section{Introduction}
\label{sec:intro}

Breast cancer ranks as the second leading cause of cancer-related deaths among women globally, making it the most prevalent cancer affecting them~\cite{american2019breast}. Early detection primarily relies on screening mammography, encompassing four images—two from each breast taken from different angles: mediolateral-oblique (MLO) from the side and Cranial-Caudal (CC) from above. 
% \textcolor{red}{Cross-view data analysis is crucial in healthcare for anomaly identification and diagnosis.}
While conventional deep learning methods for breast cancer classification have focused on single-view analysis, radiologists concurrently assess all views in mammography exams, recognizing valuable correlations that provide crucial tumor information. 
% This highlights the importance of cross-view data analysis in healthcare for identifying anomalies and making diagnoses.
% This cross-view attention approach underscores the significance of multi-view or multi-image-based Computer-Aided Diagnosis (CAD) schemes over single image-based CAD schemes. 
This highlights the importance of cross-view data analysis in healthcare for identifying anomalies and making diagnoses as well as the significance of multi-view or multi-image-based Computer-Aided Diagnosis (CAD) schemes over single image-based CAD schemes. 
Recent research in breast cancer classification and detection has applied deep learning techniques with promising outcomes. Many current studies~\cite{lopez2022hypercomplex, sun2019multi, kyono2021triage} aim to incorporate multi-view architectures inspired by radiologists' multi-view analysis, contributing to more robust and high-performing models.

Efforts to integrate multiple views into breast cancer screening through the utilization of Deep Neural Networks (DNNs) have roots in the work of Carneiro et al.~\cite{carneiro2017deep}. They trained models individually on the MLO and CC views, then employed the features from the final fully connected layer to train a multinomial logistic regression model. 
In a parallel work, Sridevi et al. in~\cite{sridevi2023combined} put forth a classification strategy utilizing CC and MLO views. Their methodology encompassed initial image normalization, pectoral muscle removal, and subsequent feature extraction through convolution and pooling layers, with the extracted features being concatenated. In \cite{khan2019multi}, Khan et al. introduced a two-stage classification strategy involving the use of extracted ROIs from four mammogram views. Various CNNs were employed as feature extractors, and the extracted features from all views were concatenated using an early fusion strategy, culminating in the final output through the classifier layer.
% In \cite{sun2019multi}, a two-view CNN was proposed to leverage complementary information from MLO and CC views, featuring multi-view and multi-scale subnetworks. The authors incorporated separate paths for each view in the multi-view subnetwork, followed by concatenation for the multi-scale subnetwork.

% work~\cite{gudhe2022multi}, the authors introduced an exploration of various fusion techniques for assessing BI-RADS findings and density. This assessment was based on four different views. Their approach involved the use of the evidential neural network algorithm, which incorporated pre-trained models and offered an interpretable method for combining features using the Dempster-Shafer evidence theory for both tasks.

% Conversely, a fundamental concept in deep learning is attention. 
Recently, attention has become a core concept in deep learning due to its widespread application.
The objective of an attention mechanism in this context is to enable the model to focus selectively on relevant local input regions and feature channels, thereby avoiding equal treatment of all locations and features. In medical image analysis, where diagnoses often hinge on specific isolated areas of concern, recent research has explored the integration of attention mechanisms into multi-view mammography analysis. However, these endeavors still partially rely on convolutional methods and do not fully exploit the inherent potential of attention mechanisms. The self-attention mechanism, intrinsic to transformers, empowers models to dynamically discern where and what to focus on, utilizing pertinent image regions or features to augment task performance. With self-attention, vision transformers~\cite{dosovitskiy2020image} excel at capturing long-range dependencies within input sequences. 

Despite the evident promise of transformers in modeling long-range dependencies, their application in multi-view mammogram analysis remains relatively uncharted territory. Some studies, such as~\cite{van2021multi}, have embraced hybrid models combining transformers and CNNs, introducing global cross-view transformer blocks to amalgamate intermediate feature maps from CC and MLO views. Another noteworthy work is~\cite{chen2022transformers}, which employed a transformer-based model for breast cancer segment detection. However, they processed multi-views at a later stage of the network, missing opportunities to capture local correlations between views and lacked results on publicly available datasets, thereby constraining comparability with existing literature.

To fully exploit multi-view insights, we present a novel transformer-based multi-view network, MV-Swin-T, built upon the Swin Transformer~\cite{liu2021swin} architecture for mammographic image classification. 
Our contributions include:

\begin{enumerate}
    \item Designing a novel multi-view network entirely based on the transformer architecture, capitalizing on the benefits of transformer operations for enhanced performance.
    \item A novel "Multi-headed Dynamic Attention Block (MDA)" with fixed and shifted window features to enable self and cross-view information fusion from both CC  and MLO views of the same breast.
    % \item Leveraging local and global correlations between views by integrating the proposed MDA modules at the early stages of the network.
    \item Addressing the challenge of effectively combining data from multiple views or modalities, especially when images may not align correctly.
    \item We present results using the publicly available CBIS-DDSM And VinDr-Mammo dataset.
\end{enumerate}
    
Moreover, to comprehend the impact of transformers and different associated modules, we introduce various architectural changes throughout the training process, analyzing their overall effects on the entire network. Detailed insights into these analyses are provided in the \ref{result} section. 
% \textcolor{red}{The goal of this paper is to investigate deeply into the transformer paradigm, offering readers a comprehensive analysis of the transformer network.}

\section{Methodology}
\label{methodology}

% In the following section, we present our proposed multi-view transformer architecture, delineate the structural variations in our models, and expound on the training protocols we have employed. Specifically,

Our approach focuses on developing a specialized network based on the Swin Transformer architecture, specifically designed for classifying unregistered multi-view mammogram pairs. 
To enhance the model's capacity, we introduce the novel Omni-Attention transformer block with advanced multi-head dynamic-attention mechanisms and both regular and shifted window configurations. It is imperative to underscore that our experimental focus is exclusively on ipsilateral-views, deliberately excluding bilateral views. Ipsilateral analysis involves diagnosis based on the CC and MLO views of the same breast, while bilateral analysis integrates findings from corresponding views of both breasts. This deliberate choice aligns with clinical norms, where radiologists typically employ ipsilateral analysis to classify breast tumor masses, and leverage symmetry information from bilateral analysis to detect asymmetries, as highlighted in~\cite{yang2021momminet}. 
% \textcolor{red}{We contend that restricting our analysis to ipsilateral views suffices for the classification of breast tumor masses, acknowledging that bilateral information remains indispensable for detecting asymmetries. }
% \textcolor{red}{Our model eschews pre-training, lacking a standard backbone architecture, and instead opts for a novel and distinctive architectural approach.}
As our architecture lacks a standard backbone structure, we avoided using pre-training and opted for a distinctive approach.

\begin{figure}[htb]
  \centering
  \centerline{\includegraphics[width= 8cm]{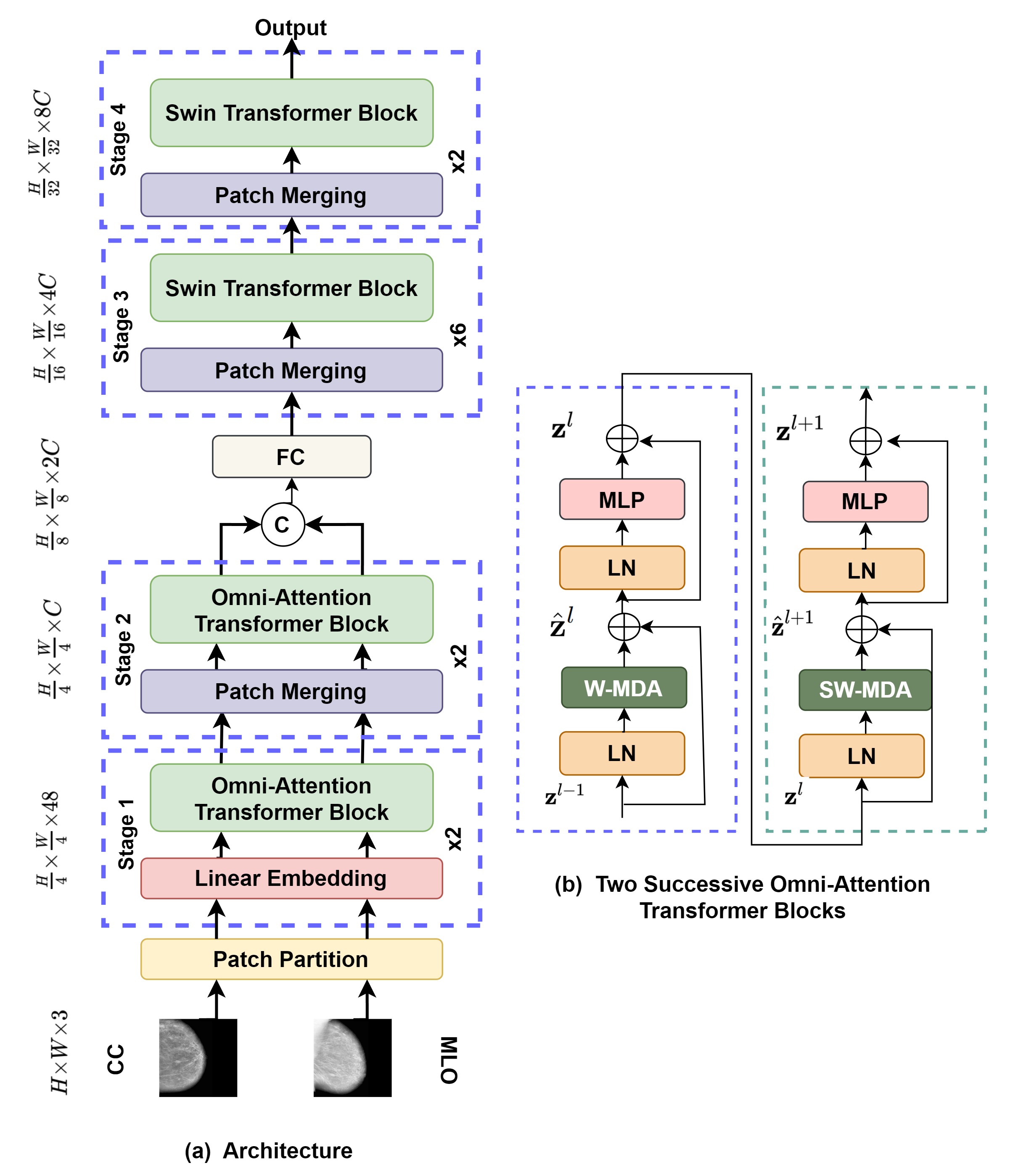}}
\caption{(a) Our proposed multi-view architecture. (b) Two successive Omni-Attention Transformer Blocks featuring W-MDA and SW-MDA components for Multi-Head Dynamic-Attention with regular and shifted window configurations. While presented as a single-input diagram for simplicity, here, Z represents the combined representations of CC and MLO, $Z = \langle Z_{CC}, Z_{MLO} \rangle.$}
\label{fig:WindowAttention}
\end{figure}

\textbf{Omni-Attention Transformer Block:} The Omni-Attention transformer block was constructed by substituting the window and shifted window multi-head self-attention (MSA) module, as presented in \cite{liu2021swin}, with the regular and shifted window multi-head dynamic attention (W-MDA and SW-MDA) module, while maintaining consistency in other layers. The architectural modification is shown in Figure~\ref{fig:WindowAttention}. 
Each multi-head dynamic attention (MDA) module is succeeded by a 2-layer Multi-Layer Perceptron (MLP) with Rectified Linear Unit (ReLU) nonlinearity in between. It is noteworthy that, contrary to the Swin Transformer paper, we have opted for ReLU nonlinearity instead of GELU nonlinearity. To enhance stability and facilitate convergence, a LayerNorm (LN) layer is applied before each MDA module and each MLP, with a residual connection applied after each module.

\begin{figure}[!t]
  \centering
  \centerline{\includegraphics[width=8.5cm]{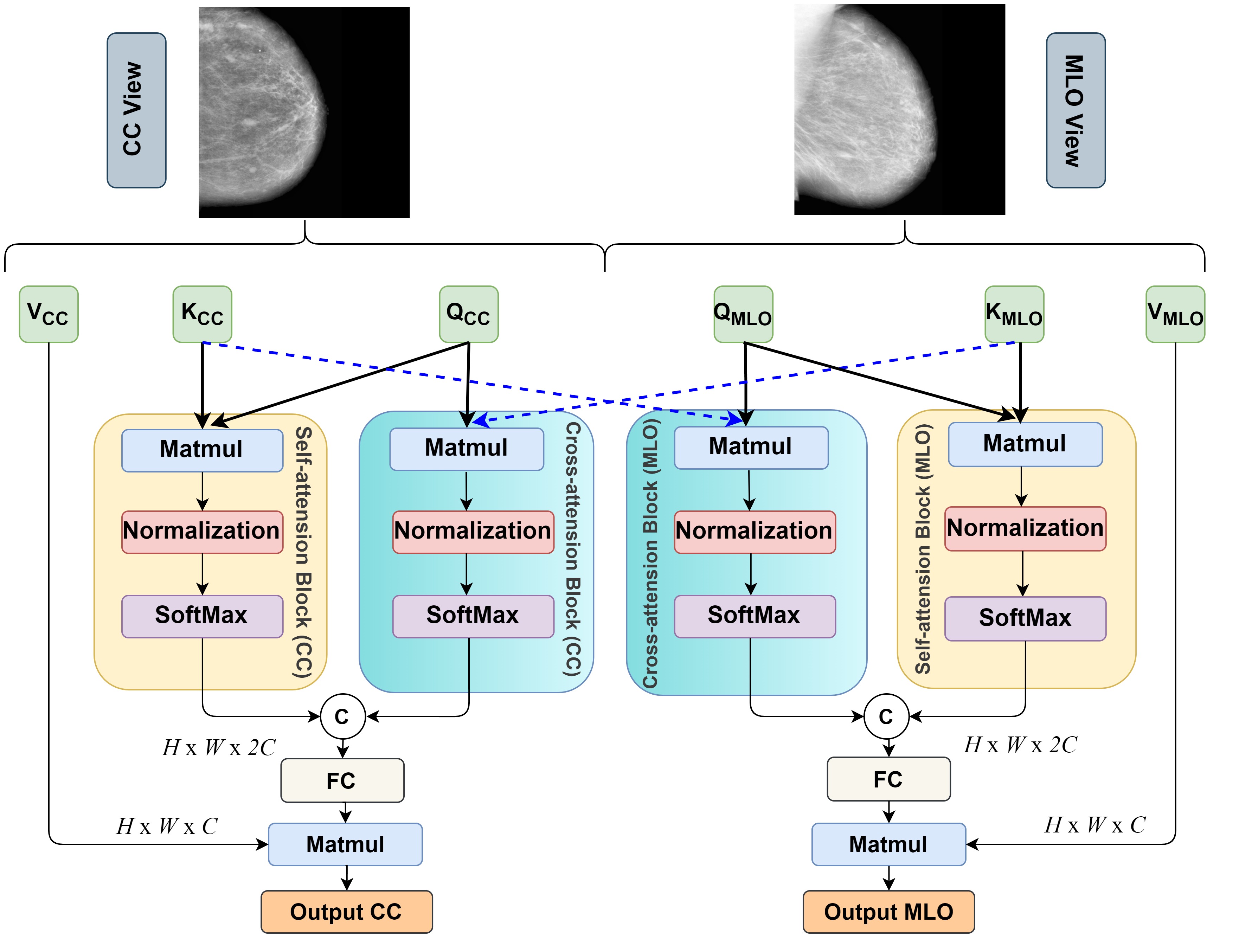}}
\caption{Illustration of self and cross attention operations within the proposed Multi-head Dynamic Attention (MDA) block for ipsilateral views. The term 'matmul' denotes matrix multiplication.}
\label{fig:AttentionMechanism}
\end{figure}

\textbf{Multi-Head Dynamic Attention:}
In each W-MDA block, we integrate both self-attention and cross-attention mechanisms for both CC and MLO views, calculated within the local windows. These windows are strategically arranged to evenly partition the image in a non-overlapping manner, with a fixed size of 7 for images with dimensions 224 $\times$ 224 and 12 for images with dimensions 384 $\times$ 384.
The calculation process for both self and cross-attention is identical, with the key distinction being that for self-attention, the $Q$, $K$, $V$ matrices are obtained from the same view. 
In contrast, for cross-view attention, the $K$ matrix is derived from the other view, while the $Q$ and $V$ matrices are derived from the origin view as showed in Figure \ref{fig:AttentionMechanism}. 
% In contrast, for cross-view attention, the $K$ matrix is derived from the other view ($\alpha$), while the $Q$ and $V$ matrices are derived from the origin view ($\beta$). 
% This cross-view attention can be denoted as $CVA_{\alpha\to \beta}$.

\begin{equation}
\scalebox{0.9}{$
\begin{aligned}
    DA_{CC} = [\text{Linear}(\sigma(\frac{Q_{CC}K_{CC}^T}{\text{scale}}) \: \oplus \sigma(\frac{Q_{CC}K_{MLO}^T}{\text{scale}}))] V_{CC}
\end{aligned}
$}
\label{eq:dacc}
\end{equation}
% \begin{equation}
% \begin{aligned}
%     DA_{CC} = [\text{Linear}(\sigma(\frac{Q_{CC}K_{CC}^T}{\text{scale}}) \: \oplus \sigma(\frac{Q_{CC}K_{MLO}^T}{\text{scale}}))] V_{CC} %\notag \\ 
% \end{aligned}
% \label{eq:dacc}
% \end{equation}
\begin{equation}
\scalebox{0.9}{$
\begin{aligned}
    W\text{-}MDA / SW\text{-}MDA = \langle DA_{CC}, DA_{MLO} \rangle % \notag
\end{aligned}
$}
\label{eq:wmda}
\end{equation}

Equation~\ref{eq:dacc} demonstrates the dynamic attention from input CC view. The $Q$, $K$, $V$ represents the query, key and value of the corresponding inputs respectively, and $\sigma$ represents softmax operation. We denote the fully connected layer as linear in the equation. $W$-$MDA$ and $SW$-$MDA$ can be constructed with corresponding dynamic attention of $DA_{CC}, DA_{MLO}$ respectively. In equation \ref{eq:equation1}, we depict the first sub-block of the Omni-Attention transformer, comprising a LayerNorm (LN) layer, a multi-head dynamic attention module (W-MDA), and a 2-layer MLP with ReLU non-linearity. Here, $l$ denotes the layer number, and $Z$ represents the feature map.

\begin{equation}
\begin{aligned}
    \hat{Z}^l &= W\text{-}MDA(LN(Z^{l-1})) + Z^{l-1}, \\
    Z^l &= MLP(LN(\hat{Z}^l)) + \hat{Z}^l.
\end{aligned}
\label{eq:equation1}
\end{equation}
\begin{equation}
\begin{aligned}
    \hat{Z}^{l+1} &= SW\text{-}MDA(LN(Z^l)) + Z^l, \\
    Z^{l+1} &= MLP(LN(\hat{Z}^{l+1})) + \hat{Z}^{l+1}.
\end{aligned}
\label{eq:equation2}
\end{equation}

\textbf{Shifted Window Multi-head Dynamic Attention:}
The dynamic-attention module based on fixed windows lacks inter-window connections, thereby limiting its modeling capabilities. To incorporate cross-window connections while preserving the computational efficiency of non-overlapping windows, we adopted a shifted window partitioning approach similar to \cite{liu2021swin}. Equation \ref{eq:equation2} illustrates the second sub-block of the Swin Transformer, comprising an LN layer, a shifted window multi-head dynamic attention module (SW-MDA), and an MLP with ReLU activation.

% Our MV-Swin-T architecture consists of Omni-Attention blocks in the first two stages of the network. After the second stages, the outputs are concatenated and passed through a fully connected layer to ensure consistency of the dimensions with a single-view mammogram. This output is then passed through the final stages of Swin Transformer blocks to get the final output.

The proposed MV-Swin-T architecture represents Omni-Attention blocks integrated into the initial two stages of the network.
Following the second stage, the outputs from different views are concatenated and channeled through a fully connected layer to maintain consistent dimensions with a single-view mammogram. 
The processed output is then directed through Swin Transformer blocks in the third and fourth stages, ultimately resulting in the final output.
% Subsequently, this processed output is passed through an identical Swin Transformer blocks of third and fourth stage; finally yieling to an output.

% The output from the second stage is concatenated, and the resulting features pass through a fully connected layer. This step ensures that the dimensions stay consistent with those of a single view. The subsequent architectural configuration closely follows that of the Swin Transformer. The concatenation and fully connected layer serve as a bridge, facilitating the seamless integration of information from the multi-view stage into the subsequent layers of the Swin Transformer.

% \[
% \begin{aligned}
% \hat{Z}^{l+1} &= SW\text{-}MDA(LN(Z^l)) + Z^l, \\
%     Z^{l+1} &= MLP(LN(\hat{Z}^{l+1})) + \hat{Z}^{l+1}.
% \end{aligned}
% \]

% (a) Our proposed multi-view architecture. (b) Two successive Omni-Attention Swin Transformer Blocks that incorporate two key components: W-MDA and SW-MDA, representing multi-head cross-attention mechanisms with both regular and shifted windowing configurations. For simplicity, we present a single-input diagram; however, it is crucial to emphasize that each block in (b) receives two inputs from the CC and MLO views. Here, Z represents the combined representations of CC and MLO, $Z = \langle Z_{CC}, Z_{MLO} \rangle.$

\section{Experiments and Results}

\subsection{Dataset \& Implementation Setting}
\label{sec:dataset}
We evaluated our proposed architecture using the Curated Breast Imaging Subset of the Digital Database for Screening Mammography (CBIS-DDSM)~\cite{lee2017curated} and VinDr-Mammo datasets \cite{nguyen2023vindr}. The CBIS dataset comprises 2462 mammography images from 1231 women, providing both CC and MLO views for most cases. After data preparation, we used 509 patient data records for training and 151 for testing, excluding cases lacking both views. Additionally, we conducted experiments incorporating calcification data from the DDSM dataset, resulting in a total of 1062 images for training and 253 for testing. The VinDr-Mammo dataset comprises 5000 four-view exams, yielding 8000 images for training and 2000 for testing when considering CC and MLO views of the same breast.

To enhance model robustness, we applied data augmentation techniques, including random horizontal and vertical flipping, as well as random rotation, to all images. Additionally, we resized all images to 224$\times$224 and 384$\times$384 before inputting them into the models. For preprocessing, we followed a similar process as discussed in \cite{sarker2022connectedunets++} for the CBIS dataset, while for the VinDr-Mammo dataset, we employed the OpenCV Python package for image normalization and enhancement. 
We categorized cases with BIRADS values of 1 to 3 as benign and those with values of 4 to 6 as malignant for the VinDr-Mammo Dataset.

We developed the model using PyTorch and conducted training using the Adam optimizer~\cite{kingma2014adam}, initializing with learning rate annealing of 0.0001 and a weight decay of 0.001. For both the dataset, we used binary cross-entropy loss and threshold value of 0.5 to make decisions about class assignments. 
%The output of a neural network for a binary classification task is usually a probability score between 0 and 1, representing the likelihood or confidence of an instance belonging to the positive class (class 1).
% Our training regimen encompassed 100 epochs with a step decay learning rate scheduler and early stopping mechanism. In all of our experiments, the best result was found in less than 50 epochs.
We conducted our training for 100 epochs, employing a ReduceLROnPlateau learning rate scheduler and an early stopping mechanism. Notably, the optimal result was consistently achieved in fewer than 50 epochs across all experiments.

\subsection{Results and Analysis}
\label{result}

\begin{table}[!t]
  \centering
  \newcommand{\hlc}{\textbf}% Highlight cell
  \caption{Comparison of the proposed architecture on different architectural settings on the CBIS-DDSM and VinDr-Mammo dataset. Here, Param.: No of parameters (in million), AUC: Area under the curve, Acc.: Accuracy, WA: Weighted Addition, Con.: Concatenation}
  \label{table1}
  \scalebox{0.85}
  {
  \begin{tabular}{*{7}{c}}
    \hline
    \multirow{2}{*}{Dataset} &  \multirow{2}{*}{\shortstack{Fusion \\ Stages}} & \multirow{2}{*}{\shortstack{Image \\ Size}} & \multirow{2}{*}{\shortstack{Attention \\ Fusion}} & Param. & AUC & Acc.\\
    & & & & & (\%) & (\%)\\
    \hline
    \multirow{5}{*}{\shortstack{CBIS \\ (Mass)}} & 4 & 224$^2$ & WA & 55 & 66.16 & 64.58 \\
    & 4 & 384$^2$ & Con. & 55 & 67.44 & 66.26 \\
    & 3 & 384$^2$ & Con. & 40 & 69.28 & 67.21 \\
    & 2 & 224$^2$ & Con. & 27 & 69.75 & 66.82 \\
    & 2 & 384$^2$ & Con. & 29 & \textbf{71.37} & \textbf{68.63} \\
    \hline
    \multirow{3}{*}{\shortstack{CBIS \\ (Mass \& \\ Calc)}} & 2 & 224$^2$ & Con. & 27 & 65.08 & 63.74 \\
    & 2 & 384$^2$ & Con. & 29 & \textbf{66.43} & \textbf{65.37} \\
    & 2 & 448$^2$ & Con. & 30 & 64.66 & 63.19 \\
    \hline
    \multirow{2}{*}{\shortstack{VinDr- \\ Mammo}} &  2 & 224$^2$ & Con. & 27 & 95.88 & 95.00 \\
    & 2 & 384$^2$ & Con. & 29 & \textbf{96.08} & \textbf{95.50} \\
    \hline
  \end{tabular}
  }
\end{table}

We explored various configurations, and the optimal model, featuring cross-view attention, is illustrated in Figure~\ref{fig:WindowAttention}. Table~\ref{table1} presents a comparative analysis of different architectural settings, specifically focusing on the 'Fusion Stage' column, which signifies CC and MLO view fusion at the stage level. Our experiments revealed that implementing fusion after the second stage consistently yielded the best results across different datasets. This reduction in parameters significantly enhances model efficiency, preventing overfitting and reducing computational complexity, particularly beneficial when handling a limited number of training images. 
Also, early integration after the second stage ensures a comprehensive representation, considering both views, thereby improving the model's contextual understanding and overall performance.

The "Attention Fusion" column indicates how attention from CC and MLO views is combined in the MDA modules. For this, we conducted experiments with weighted addition and concatenation. In the weighted addition method, we tested various weight values ranging from 0.9 to 0.5. However, we only presented results for the value set of 0.9 and 0.1, where 0.9 is multiplied with the same view, and 0.1 is multiplied with the cross view, as this setting yielded the best result among all the value sets. Nevertheless, the overall best result was achieved with the concatenation method.
% Initially, we explored the use of two distinct networks for CC and MLO views, collectively referred to as a two-tower network. This architecture resulted in a parameter count exceeding twice that of a standard Swin-Transformer. Furthermore, we implemented W-MDA and SW-MDA within the two-tower network.
% Results obtained from the W-MDA blocks guided subsequent experiments, wherein we investigated operations between cross-attention and self-attention values. We explored element-wise addition and a gating block of the self and cross-attention blocks, leading to increased parameters and decreased accuracy. Notably, the superior performance was achieved through the concatenation of these two blocks, as illustrated in Figure 1. To manage the dimensionality of the concatenated block, we passed its value through a fully-connected layer.
Additionally, we examined two image sizes (224$\times$224 and 384$\times$384) and diversified the training datasets to include mass and mass-calc. The model delivers better result for image size 384$\times$384. 
% We conclude that larger image size provides more pixels for the model to extract features from benefitting from an increased receptive field, potentially leading to more representative and discriminative features. 
% \textcolor{red}{We achieved the best result with an image size of 384$\times$384 which is bold in Table~\ref{table1}.}
We conclude that as the success of the classification task relies not only on accurate mass localization but also on effectively attending to the mass boundary, increasing the image size augments the network's ability to comprehend mass boundaries.

% Following a comprehensive exploration, the most effective network configuration emerged as one utilizing W-MDA and SW-MDA for the initial two stages. The outputs of these networks were concatenated, processed through a fully-connected layer, and subsequently passed through two stages of regular Swin-Transformer blocks to obtain the final output. 
%This novel architecture is referred to as the fused-two-tower network.

\begin{table}[!t]
  \centering
  \newcommand{\hlc}{\textbf}% Highlight cell
  \caption{Comparing our MV-Swin-T to baseline Swin Transformer (Swin-T) on CBIS-DDSM and VinDr-Mammo datasets, using multi-view input versus single views. Results presented for an image size of 384$\times$384.}
  \label{table2}
  \scalebox{0.85}
  {
  \begin{tabular}{*{5}{c}}
    \hline
    \multirow{2}{*}{Dataset} &  \multirow{2}{*}{\shortstack{Architecture \\ Name}} & \multirow{2}{*}{\shortstack{Number \\ of View}} & AUC & Acc.\\
    & & & (\%) & (\%)\\
    \hline
    \multirow{2}{*}{\shortstack{CBIS \\ (Mass)}} & Swin-T & Single & 60.26 & 59.64 \\
    & MV-Swin-T (ours) & Two & \textbf{71.37} & \textbf{68.63} \\
    \hline
    \multirow{2}{*}{\shortstack{CBIS \\ (Mass \& Calc)}} & Swin-T & Single & 61.28 & 60.46 \\
    & MV-Swin-T (ours) & Two & \textbf{66.43} & \textbf{65.37} \\
    \hline
    \multirow{2}{*}{\shortstack{VinDr- \\ Mammo}} &  Swin-T & Single & 95.88 & \textbf{95.50} \\
    & MV-Swin-T (ours) & Two & \textbf{96.08} & \textbf{95.50} \\
    \hline
  \end{tabular}
  }
\end{table}

Table \ref{table2} presents a comparative analysis of our proposed MV-Swin-T architecture and the baseline Swin-T architecture, operating on single views. The results distinctly showcase the superior performance of our proposed model across all datasets. This further bloster the notion that assessment based on multi-view fusion can effectively integrate
relevant information from both views, mitigating noise interference from single views and yielding more accurate predictions. The model particularly showed optimum results for VinDr-Mammo dataset.
We attribute this performance gap to the varying image quality between the two datasets. Additionally, the larger size of the VinDr dataset could contribute to the superior performance, as deep learning models often benefit from larger datasets for optimal results.

\section{Discussion and Conclusion}
\label{conclusion}

Transformers have gained prominence as effective architectures for capturing long-range dependencies in recent years. However, their potential in the field of multi-view mammogram analysis remains significantly untapped. In this research, we present a pure transformer-based multi-view network for breast cancer classification, designed to emulate the way radiologists interpret multi-view mammograms and utilize the information present in ipsilateral views. 
We introduce a novel Multi-Headed Dynamic Attention (MDA) block that implicitly learns to emphasize cancer-related local anomalies and accentuates essential features by exploring cross-view information between two mammogram views. This study involved extensive experimentation, exploring various configurations across multiple datasets. 
% The comparison of these configurations revealed valuable insights. 
In the concluding section, we showcased the performance of our multi-view approach in contrast to single-view methods. The results underscored the superiority of the multi-view approach, highlighting its effectiveness in enhancing overall performance.  We believe that our approach represents a significant breakthrough in this research field, opening the door to novel techniques that can process medical imaging exams more similar to how radiologists analyze.
Exploring strategies to investigate the scalability of the multi-view approach to even larger datasets could provide additional insights and improvements to the model's generalization capabilities.

\section{Acknowledgments}
\label{sec:acknowledgments}
This work has received support from the following grants and awards.
\begin{itemize}
  \item National Science Foundation under Grant No. OIA- 2148788.
  \item Nevada DRIVE program (Doctoral Research in Innovation, Vision and Excellence)
\end{itemize}

\section{Compliance with Ethical Standards}
This research study was conducted retrospectively using human subject data made available in open access by \cite{lee2017curated, nguyen2023vindr}. Ethical approval was not required as confirmed by the license attached with the open access data.

\bibliographystyle{IEEEbib}
\bibliography{strings,refs}

\end{document}